\documentclass[conference]{IEEEtran}
\usepackage{times}

\usepackage[numbers]{natbib}
\usepackage{multicol}
\usepackage[bookmarks=true]{hyperref}

\usepackage{times}
\usepackage{dsfont}
\hypersetup{
    colorlinks=true,
    linkcolor=blue,
    filecolor=magenta,      
    urlcolor=cyan,
    pdftitle={Overleaf Example},
    pdfpagemode=FullScreen,
    }
\usepackage[utf8]{inputenc} 
\usepackage[T1]{fontenc}    
\usepackage{url}            
\usepackage{booktabs}       
\usepackage{amsfonts,bm}       
\usepackage{amssymb}
\usepackage{float}
\usepackage{nicefrac}       
\usepackage{microtype}      
\usepackage{xcolor}         
\usepackage{bbm}

\usepackage{subcaption}
\usepackage{amsmath,amssymb}

\usepackage{lipsum}
\usepackage{algorithm}
\usepackage{algpseudocode}
\usepackage{graphicx}
\usepackage{framed}

\usepackage{siunitx}

\usepackage{enumitem}

\usepackage{mathtools}
\graphicspath{ {images/} }

\usepackage{multicol}
\usepackage{svg}
\usepackage{multirow}
\usepackage{diagbox}
\usepackage[normalem]{ulem}

\newcommand{\va}{{\bf a}}

\newcommand{\vo}{{\bf o}}
\newcommand{\vp}{{\bf p}}
\newcommand{\vs}{{\bf s}}

\newcommand{\vz}{{\bf z}}

\newcommand{\vx}{{\bf x}}

\newcommand{\vtheta}{\boldsymbol{\theta}}

\begin{document}

\title{Implicit Neural-Representation Learning for \\
Elastic Deformable-Object Manipulations}


\author{
\IEEEauthorblockN{Minseok Song}
\IEEEauthorblockA{KAIST\\
hjmngb@kaist.ac.kr
\vspace*{-1\baselineskip}}
\and
\IEEEauthorblockN{JeongHo Ha}
\IEEEauthorblockA{KAIST\\
hajeongho95@kaist.ac.kr
\vspace*{-1\baselineskip}}
\and
\IEEEauthorblockN{Bonggyeong Park}
\IEEEauthorblockA{KAIST\\
iampbk@kaist.ac.kr
\vspace*{-1\baselineskip}}
\and
\IEEEauthorblockN{Daehyung Park}
\IEEEauthorblockA{KAIST\\
daehyung@kaist.ac.kr
\vspace*{-1\baselineskip}}
}

\maketitle

\begin{abstract}
We aim to solve the problem of manipulating deformable objects, particularly elastic bands, in real-world scenarios. However, deformable object manipulation (DOM) requires a policy that works on a large state space due to the unlimited degree of freedom (DoF) of deformable objects. Further, their dense but partial observations (e.g., images or point clouds) may increase the sampling complexity and uncertainty in policy learning. To figure it out, we propose a novel implicit neural-representation (INR) learning for elastic DOMs, called INR-DOM. Our method learns consistent state representations associated with partially observable elastic objects reconstructing a complete and implicit surface represented as a signed distance function. Furthermore, we perform exploratory representation fine-tuning through reinforcement learning (RL) that enables RL algorithms to effectively learn exploitable representations while efficiently obtaining a DOM policy. We perform quantitative and qualitative analyses building three simulated environments and real-world manipulation studies with a Franka Emika Panda arm. Videos are available at \url{http://inr-dom.github.io}.

\end{abstract}

\IEEEpeerreviewmaketitle

\section{Introduction}

\begin{figure}[t]
 \centering
  \includegraphics[width=\columnwidth]{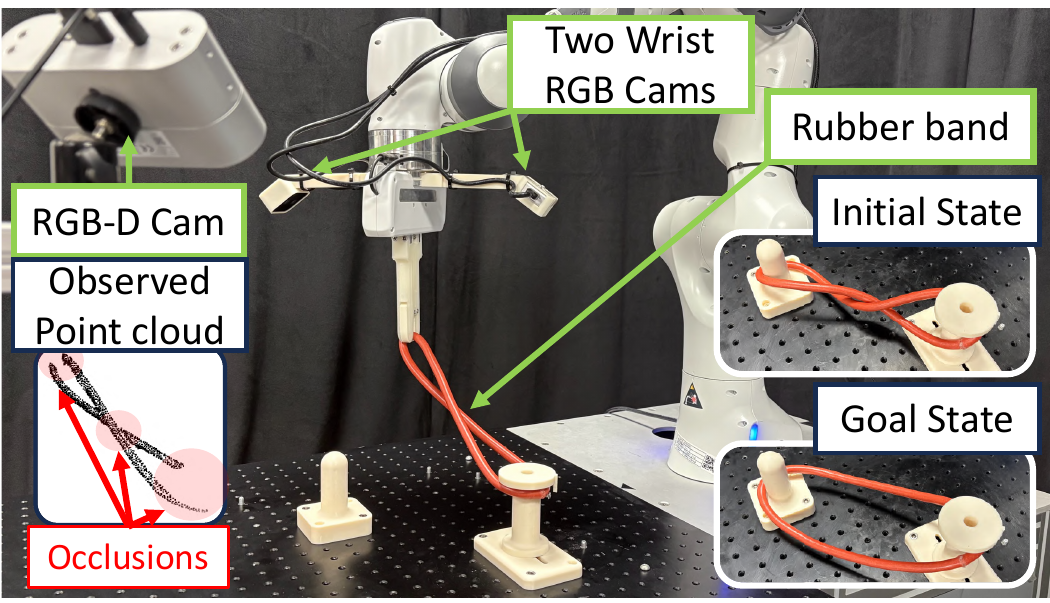}
  \caption{A capture of deformable object manipulation task that requires  disentangling elastic bands between two poles. The deformable and stretchable nature of bands increases the complexity of state representation. Further, the $360^\circ$ twists create self-occlusions, significantly reducing the consistency of state embeddings. Our INR-DOM effectively captures the occlusion-robust implicit representation of the bands and efficiently generates real-world applicable manipulation policies. 
  }
  \vspace{-1em}
  \label{fig:main}
\end{figure}

Deformable object manipulation (DOM), as shown in Fig.~\ref{fig:main}, presents a major challenge in automation and has attracted increasing attention in robotics over the past decade~\cite{gu2023survey}. Researchers have investigated a variety of DOM tasks, such as grasping~\cite{wang2018knot}, folding~\cite{salhotra2022learning}, wearing~\cite{ren2023autonomous}, threading~\cite{mitrano2024grasp}, winding~\cite{murase2017kullback}, tangling~\cite{shivakumar2023sgtm}, and bagging~\cite{slipbagging2023}. These tasks introduce challenges in the domains of perception, modeling, planning, and control~\cite{zhu2022challenges}, due to infinite degrees of freedom (DoF) and nonlinear interaction dynamics of deformable objects (DO). Further, dense but partial observations---often resulting from self-occlusions---also increase sampling complexity and uncertainty in policy learning. 

In DOM, data-driven modeling approaches are increasingly gaining attention with their extensive representation capabilities for downstream tasks. For example, Lippi et al. reduce a high-dimensional space of DOs into a low-dimensional state graph to facilitate planning~\cite{lippi2020latent}. Researchers often model a DO as a particle-based interaction graph to describe detailed topological structures~\cite{li2019learning, longhini2023edo}. Recent studies aim to capture comprehensive graph-based models by reconstructing complete geometries such as point clouds and meshes from partial observations~\cite{huang2022mesh}. However, the discrete nature of particle- or graph-based models often fails to consistently represent the flexible and smooth surfaces of DOs.

Meanwhile, signed distance fields (SDFs) are receiving increasing interest with their ability to represent complex, non-convex geometries of objects~\cite{zheng2022sdf}. SDFs not only describe the surfaces of objects, but also provide spatial information at a distance, making them well suited to depict physical interactions and facilitating manipulation planning~\cite{zucker2013chomp}. To capture continuous and fine details, researchers parameterize SDFs using neural networks (i.e., implicit signed distance function)~\cite{Sitzmann_2020, David2024cloth}. However, most studies that involve SDFs in DOM focus primarily on enhancing encoding for improved reconstruction~\cite{wi2022virdo}, which may significantly distract the exploration process of reinforcement learning (RL)-based policy learning.

We propose a novel implicit neural representation (INR) learning method for elastic DOM, which we call INR-DOM. Our method focuses on learning consistent and occlusion-tolerant representations of partially observable DOs and enhances task-relevant representations to optimize their manipulation policies efficiently. INR-DOM incorporates a two-stage learning process with two types of losses: $1$) pre-training utilizes reconstruction and regularization losses to develop an occlusion-robust yet dense representation encoder, suitable for stretched or intertwined DOs; $2$) fine-tuning employs contrastive learning and RL losses to refine representations, boosting exploitability and effectiveness in policy learning. The use of implicit SDF during pre-training allows for the reconstruction of complete and implicit surfaces of highly elastic DOs. The integration of contrastive loss during fine-tuning further enables the encoder to effectively identify and handle complex, enclosed states, such as twisted rubber bands. We particularly introduce a temporal- and instance-wise key assignment method for time-series contrastive learning to better represent correlations between similar manipulation sequences.

We conduct quantitative and qualitative studies in both simulated and real-world manipulation environments. By constructing a 3D shape-recovery benchmark of nine elastic rubber bands, we show INR-DOM's consistent and occlusion-robust representation capabilities. Applying them to simulated environments, we then demonstrate the superior state-representation and policy-learning capabilities across three simulated environments (i.e., \textit{sealing}, \textit{installation}, and \textit{disentanglement} tasks). INR-DOM excels in accurately recovering complete geometries from partial observations, surpassing state-of-the-art baseline methods, particularly in handling stretched or intricately enclosed states of rubber bands. Further, the fine-tuned representation captures both task-relevant and -irrelevant details, facilitating efficient policy convergence in RL and leading to significantly higher task success rates up to $41\%$ higher than the next-best baseline approach. We also demonstrate the real-world applicability of INR-DOM with a Franka Emika Panda robot.

In summary, key contributions of this paper are threefold:
\begin{itemize}[leftmargin=*]    
    \item We introduce an implicit neural-representation learning method that provides consistent and occlusion-robust representations from visual observation of deformable objects.
    \item We develop an effective representation-refining approach using a contrastive loss to capture task-relevant state information for RL.
    \item We demonstrate that INR-DOM significantly improves convergence stability in policy learning and success rate for DOM tasks, in both simulations and real-world settings.
\end{itemize}
\section{Related Works}
\label{sec:related}
We investigate representation models and their learning or update methodologies for DOM tasks. 

\noindent\textbf{Representation models}:
The expressiveness of representations stems from the capacity of the underlying models. Most model-based approaches represent deformations using discrete structures composed of a finite number of elements, such as points or lines. Earlier approaches, including the finite element method (FEM), approximate complex, irregular geometries by partitioning objects into smaller elements that collectively represent the overall shape~\cite{lin2015soft, petit2017tracking}. However, their reliance on the mesh structures limits their ability to handle large deformations, particularly in elastic materials (e.g., bands). Alternatively, researchers often adopt particle-based, data-driven representations: point clouds~\cite{chen2024differentiable} and voxels~\cite{duisterhof2024deformgs}. However, these approaches often struggle to capture continuum behaviors such as stretching or compression. Recently, studies build interaction graphs to leverage structural connectivity through graph-based representations~\cite{li2019learning, lin2022learning, longhini2023edo}. Nonetheless, the inherently discrete nature of these models gives challenges representing continuous surfaces.

Alternatively, researchers often adopt model-free approaches that directly map raw observations to feature vectors\textemdash using 2D convolutional encoders for RGB images~\cite{yan2021learning, salhotra2022learning} or PointNet~\cite{qi2017pointnet} for 3D point clouds~\cite{li2024deformnet}. However, these methods lack the dense geometric representations required for downstream tasks such as precise tying or untwisting. To address this limitation, implicit neural representations have demonstrated superior capabilities in capturing dense correspondence and dynamics in deformable objects. Shen et al. reconstruct high-fidelity voxel geometries from simulated RGB-D images~\cite{shen2022acid}, while Wi et al. model the continuous representation of deformed geometries based on external forces and their locations~\cite{wi2022virdo}. Our approach not only provides dense descriptors but also distinguishes complex and overlapping shapes in real-world scenarios.

\noindent\textbf{Representation Learning}: Early approaches often rely on autoencoders with reconstruction losses to obtain low-dimensional features. While this enable training with large unlabeled datasets, the resulting representations are task agnostic and may distract explorations, leading to sample inefficiency in RL~\cite{zhang2021learning}. To address this issue, researchers introduce end-to-end learning approaches that jointly train the encoder and the action network~\cite{watter2015embed}. In addition, contrastive learning~\cite{hadsell2006dimensionality} have emerged as another metric-learning technique for representation learning by promoting similarity constraints on the same or nearby data while maximizing separation from unrelated one~\cite{shen2022acid}. Leveraging this contrastive learning, Srinivas et al. improve the sample efficiency in model-free RL~\cite{laskin2020curl} by integrating the contrastive loss as an auxiliary objective. However, the standard separation of values in time series often makes it difficult to preserve similarities between temporally correlated time-series instances in self-supervised learning~\cite{lee2024soft}. In this work, we fine-tune the encoder using a contrastive loss and enhance the sample selection strategy by exploiting temporal structure in experienced trajectories, aiming to improve exploration efficiency and promote more generalizable latent representations.

\begin{figure*}[t]
    \centering
    \includegraphics[width=1\textwidth]{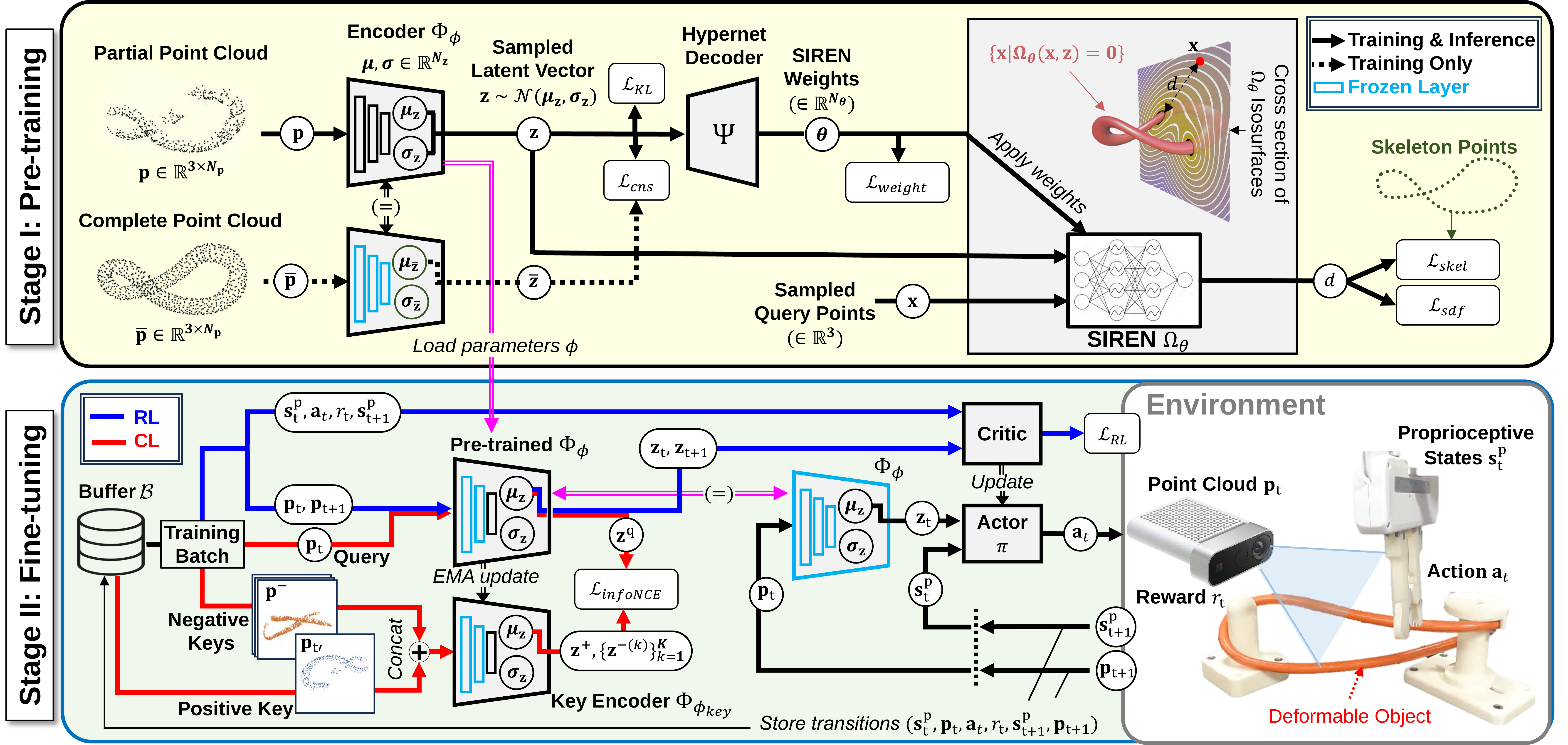}
    \caption{\label{fig:DORing_framework} An overview of INR-DOM framework that aims to train the occlusion-robust state representation encoder $\Phi_\phi$, parameterized by $\phi$, of deformable objects (DOs) as well as the manipulation policy $\pi$. The training framework consists of two stages: 1) The first stage pre-trains a PointNet-based partial-to-complete variational autoencoder $(\Phi_\phi, \Psi)$ that embeds a partial point cloud $\vp$ of a target DO into a latent embedding $\vz$ and recovers the parameters $\vtheta$ of an implicit signed distance field (SDF) network $\Omega_\theta$. This stage predicts full geometries leveraging two loss functions $\mathcal{L}_{\text{SDF}}$, $\mathcal{L}_{\text{skel}}$, along with three regularization loss functions: $\mathcal{L}_{\text{KL}}$, $\mathcal{L}_{\text{weight}}$, and $\mathcal{L}_{\text{cns}}$. 2) The second stage then improves the task-relevant representation power of the encoder $\Phi_\phi$ by jointly optimizing reinforcement learning (blue) with the loss $\mathcal{L}_{RL}$ and the contrastive learning (red) with the loss $\mathcal{L}_{\text{infoNCE}}$.}   
\end{figure*}
\section{Methodology}
\label{sec:methods}

\subsection{Overview}
The objective of INR-DOM is to jointly learn consistent state embeddings ($\vs\in\mathcal{S}$) from partial observations ($\vo\in\mathcal{O}$) and an RL policy $\pi$ for DOM. We formulate the policy learning problem as an MDP tuple $<\mathcal{S},\mathcal{A},T,R>$ with state space $\mathcal{S}$, action space $\mathcal{A}$, stochastic transition function $T$, and reward function $R$. The goal of the MDP is to find an optimal policy $\pi^* : \mathcal{S} \rightarrow \mathcal{A}$ that maximizes the cumulative discounted reward. Fig.~\ref{fig:DORing_framework} shows the overall architecture for the state-representation learning (SRL) integrated with policy learning. The input observation $\vo$ is a combination of proprioception and exteroception information, including a point cloud $\vp \in \mathcal{P}$ representing DOs, where $\mathcal{P}$ denotes the space of observable 3D point clouds. We derive a coherent, low-dimensional state $\vs$ by embedding the porint cloud $\vp$ into a latent vector $\vz \in \mathcal{Z}$ through a neural SRL function.

We design the neural SRL function, $\Phi: \mathcal{P}\rightarrow \mathcal{Z}$, as an encoder that embeds not only consistent but also task-relevant information of DOs from partial observations. To embed state representations for DOM policy learning, we introduce a two-stage training framework; Stage I) reconstruction-based pre-training (see details in Sec.~\ref{subsec:Reconstruction-based pre-training}) and Stage II) contrastive-based fine-tuning (see details in Sec.~\ref{subsec:Contrastive-based fine-tuning}). Note that this work considers training in a physics simulation and transfers the learned SRL function to the real world. The following subsections detail the architecture and training procedures.

\subsection{Reconstruction-based pre-training} \label{subsec:Reconstruction-based pre-training}
The goal of the pre-training stage is to learn an occlusion-tolerant yet dense representation $\vz$ from a partial point cloud $\vp$ of elastic DOs. A key challenge lies in learning distinguishable representations of stretched or intertwined DOs. To address this, we detail the proposed SRL network architecture and its training process with the associated loss functions. 

\noindent\textbf{Network architecture}. We design a reconstruction-based partial-to-complete variational autoencoder (VAE), capable of generating the parameters ${\vtheta} \in \Theta$ of an implicit SDF network. Our proposed network consists of a PointNet-based encoder $\Phi$ parameterized by $\phi$, a hypernetwork decoder $\Psi$, and an implicit neural SDF network $\Omega$ parameterized by $\vtheta$, 
\begin{align}
\Phi_\phi: \mathcal{P}\rightarrow \mathcal{Z},\quad \Psi: \mathcal{Z}\rightarrow \Theta,\quad \Omega_{\vtheta}: \mathcal{X} , \mathcal{Z}\rightarrow \mathbb{R} ,
\end{align}
where $\mathcal{Z}$, $\Theta$, and $\mathcal{X}$ are the spaces of state embeddings, implicit network parameters, and Cartesian points ($\in \mathbb{R}^3$), respectively. 

The encoder $\Phi$ is a modified PointNet that takes a point cloud $\vp$ of a target DO, observed from a depth image, and returns a low-dimensional latent vector $\vz\in\mathcal{Z}$. During training, we sample $\vz$ using the reparameterization trick, while during testing, we use the predicted mean of a Gaussian distribution. The modification is the removal of T-Net in the original PointNet to distinguish the translation and rotation of point clouds. The vector serves as a global representation of the complete geometry, containing its continuous surface and kinematic structure information, despite the incomplete and occluded nature of the input point cloud. In this work, we use $\mathcal{Z} \subset \mathbb{R}^{N_\vz}$ and $N_\vz=64$.

Unlike conventional VAE, the decoder $\Psi$ is a hypernetwork that takes the latent vector $\vz$ and generates weights $\vtheta$ for the implicit SDF network $\Omega$. The network is a multi-layer perceptron (MLP) with three hidden layers, where each layer has $256$ units with ReLU activations. 

The implicit SDF network $\Omega_{\vtheta}$ is a variant of SIREN Hyponetwork~\cite{Sitzmann_2020}, parameterized by the decoded weights $\vtheta$, which takes a Cartesian query coordinate $\vx$ and a current latent vector $\vz$. The network then returns a corresponding signed distance $d$ from the nearest object surface, where the positive sign indicates outside of the surface and the negative sign indicates inside. The nature of the implicit network allows for retrieving continuous and differentiable distances from a complete surface. In contrast to the original SIREN, we input the point-cloud embedding $\vz$, which helps the decoder $\Psi$ effectively learn the residual part similar to DeepSDF~\cite{Park_2019}. Our network consists of a three-layer MLP with $32$ units per layer and sinusoidal activations.

\noindent\textbf{Losses}. We introduce an INR learning loss to train the proposed network through simulated DOM data and transfer the learned model to the real world. Our proposed loss $\mathcal{L}$ is a linearly weighted combination of shape and structure reconstruction losses as well as regularization losses,
\begin{align}
    \mathcal{L}=\underbrace{\mathcal{L}_{\text{SDF}} +  \lambda_1 \mathcal{L}_{\text{skel}} }_{\text{reconstruction}} + \underbrace{ \lambda_2 \mathcal{L}_{\text{KL}} + \lambda_3 \mathcal{L}_{\text{weight}} + \lambda_4 \mathcal{L}_{\text{cns}} }_{ \text{regularization} } ,
\end{align}
where $\lambda_i$ are non-negative constants ($i\in[1,4]$). 

First of all, the reconstruction-based losses aim to complete a given partial point cloud while precisely representing its continuous and dense geometry. We introduce two loss functions:
\begin{enumerate}[leftmargin=*]
\item $\mathcal{L}_{\text{SDF}}$: An SDF loss to precisely fit the generated SDF with a ground-truth complete point-cloud $\bar{\mathcal{P}}$. Let $\mathcal{Q}\in\mathbb{R}^3$ and $\bar{\mathcal{Q}}$ denote all sampled query points and their subset on the object surface, respectively. Based on the Eq.~(6) of \cite{Sitzmann_2020}, we define the loss function as:
\begin{align}
             \mathcal{L}_{\text{SDF}} &= \int_{\vx\in\mathcal{Q}} |\left\| \nabla_{\mathbf{x}} \Omega_{\vtheta} (\mathbf{x}, \vz) \right\| - 1 | \; d\mathbf{x} \nonumber \\
             &+ \int_{\vx\in\bar{\mathcal{Q}}} | {\Omega_{\vtheta} (\mathbf{x}, \vz)} | + \left( 1 - \nabla_{\mathbf{x}} \Omega_{\vtheta} (\mathbf{x}, \vz) \cdot \mathbf{n}(\mathbf{x})\right) d\mathbf{x} \nonumber \\
        &+ \int_{\vx\in\mathcal{Q} \setminus \bar{\mathcal{Q}}} \exp \left( -\alpha \cdot | {\Omega_{\vtheta} (\mathbf{x}, \vz)}| \right) d\mathbf{x},   
        \label{eq_loss_sdf}
\end{align}
where $\mathbf{n}(\vx)$ is the surface normal vector at the query point $\vx$, $\alpha$ is a constant, and $\vz$ is the current latent vector. The first term in Eq.~(\ref{eq_loss_sdf}) constrains inconsistent SDF gradients since the gradients are mostly one, except on critical points such as medial-axis points. The second term penalizes when the on-surface points have non-zero distances and also their SDF gradient direction is not aligned with the ground-truth normal vector. Lastly, the third term regularizes off-surface points not to have large values. Note that we obtain $\mathbf{n}(\vx)$ by simulation at each time step. We also sample not only random points for $\mathcal{Q}$ but also near-surface points to accurately estimate the distance around the surface.
\item $\mathcal{L}_{\text{skel}}$: A skeleton loss that is the measure of how far the estimated medial-axis point of a DO is off from the ground-truth medial axes, where the medial-axis point exhibits the same distance to multiple boundaries (i.e., surface) on the object ~\cite{SAHA20173_skel}. This is crucial to accurately recover the geometries of intertwined or occluded regions in a partial point cloud. Here, we formulate the loss as follows: 
\begin{align}
    \mathcal{L}_{\text{skel}} = \int_{\ \vx\in \mathcal{Q}_*  }\log\Bigl( \max (\Delta\Omega_{\vtheta}(\mathbf{x}, \vz), \epsilon)^{-1}\Bigl)d\vx,
\end{align}
where $Q_*$ are the ground-truth medial-axis points, $\Delta\Omega_{\vtheta}(\vx, \vz)$ represents the Laplacian of the query point $\vx$ given $\Omega_{\vtheta}$, and $\epsilon$ means a small constant introduced to prevent division by zero. Around the medial-axis points, $\Delta\Omega_{\vtheta}(\vx,\vz)\rightarrow \infty$, as these points correspond to the local minima of $\Omega_{\vtheta}$. 
\end{enumerate}

Next, the regularization losses aim to constrain the divergence of latent vector space and weights. We introduce three loss functions:
\begin{enumerate}[leftmargin=*]
\item $\mathcal{L}_{\text{KL}}$: A Kullback-Leiber divergence loss (i.e., a regularization loss) that is a negative divergence $-D_{KL}(\Phi(\vz|\vp) \| p(\vz))$ from a prior $p(\vz)$, parameterized by $p_0(\vz)=\mathcal{N}(0,1)$, to the variational approximation $\Phi(\vz | \vp)$ of $p(\vz | \vp)$. Our encoder $\Phi$ estimates the mean $\mu \in \mathbb{R}^{N_\vz}$ and standard deviation $\sigma \in \mathbb{R}^{N_\vz}$ of the posterior distribution $\Phi(\vz|\vp)$. Then, $\mathcal{L}_{\text{KL}}=-0.5\cdot \left( \log(\sigma^2) + 1 - \mu^2 - \sigma^2 \right)$.

\item $\mathcal{L}_{\text{weight}}$: A weight-regularization loss defined as $\frac{1}{N_{\vtheta}}\|\vtheta \|^2_2$, where $N_{\vtheta}$ is the number of weight parameters including biases, adopted from \cite{Sitzmann_2020}. This formulation promotes a low-frequency SDF solution. In this work, we use $N_{\vtheta}=41,857$, which includes the parameters of the original SIREN along with our modifications.
\item $\mathcal{L}_{\text{cns}}$: A consistency loss, defined as $\frac{1}{N_\vz}\| \vz - \bar{\vz} \|^2_2$, where $\vz$ and $\bar{\vz}$ are the embeddings of the partial point cloud $\vp$ and the complete point cloud $\bar{\vp}$, respectively. This loss regularizes the embedding of the partial point cloud $\vp$ to be close to that of the complete point cloud $\bar{\vp}$. Therefore, INR-DOM enables various views of point clouds to be mapped onto the same region in the latent space. 
\end{enumerate}

\subsection{Policy learning with fine-tuning} \label{subsec:Contrastive-based fine-tuning}

The objective of this fine-tuning stage is to learn sample-efficient representations while optimizing a DOM policy. To address this, we combine RL and contrastive learning, similar to \cite{laskin2020curl}. Fig.~\ref{fig:DORing_framework} Stage II shows the overall architecture that updates the pre-trained encoder $\Phi_{\phi}$, parameterized by $\phi$, through RL while contrasting experienced sequences in the replay buffer $\mathcal{B}$. To better represent correlations between similar sequences, we introduce temporal- and instance-wise representation selections (i.e., key assignments). In the following, we detail our proposed RL-based fine-tuning and contrastive learning methods.

\noindent\textbf{Reinforcement learning}. We apply a deep RL framework to DOM as shown in Fig.~\ref{fig:DORing_framework} Stage II. Our framework aims to manipulate a DO, observed as a point cloud $\mathbf{p}_t$ at time step $t$, to a desired point-cloud status $\mathbf{p}_{\text{des}}$ using a six degree-of-freedom (DoF) robotic arm. We assume an RGB-D camera is available as shown in Fig.~\ref{fig:DORing_framework}. Below, we describe our MDP details with the RL-based loss function $\mathcal{L}_{\text{RL}}$ to update the encoder $\Phi_\phi$: \label{textbf:Reinforcement learning}
\begin{itemize}[leftmargin=*]
\item \textbf{State}: We define the state $\mathbf{s}_t$ as a tuple $(\mathbf{s}_t^{\text{p}},\mathbf{p}_t)$ that consists of the arm and DO state information, where $\mathbf{s}_t^{\text{p}}$ is a proprioceptive state vector $[\mathbf{x}_{\text{ee}}, \mathbf{q}_{\text{ee}}, \dot{\mathbf{x}}_{\text{ee}}, \dot{\mathbf{q}}_{\text{ee}}, s_{\text{gripper}}]$ of the arm end-effector, $s_{\text{gripper}}$ is the arm gripper's open and close state ($\in\{0,1\})$.
\item \textbf{Action}: We define the action $\mathbf{a}_t$ as a tuple $(\mathbf{a}_t^{\text{lin}}, \mathbf{a}_t^{\text{ang}}, \mathbf{a}_t^{\text{gripper}})\in\mathbb{R}^7$, where $\mathbf{a}^{\text{lin}}_t\in\mathbb{R}^3$ and $\mathbf{a}^{\text{ang}}_t\in\mathbb{R}^3$ are the linear and angular velocities of the end effector, respectively. $\mathbf{a}_t^{\text{gripper}}\in\{0,1\}$ indicates the gripper's open and close action.
\item \textbf{Reward}: We define the reward function $R$ as a linear combination of sparse and dense rewards. Let $\textrm{CD}(\cdot,\cdot)$ be a function that returns the Chamfer distance between two point clouds. Then, 
\begin{align}
    \hspace{-0.3em}R(\vs_t, \va_t)= &+100\cdot \mathbbm{1}_{\textrm{CD}(\mathbf{p}_t, \mathbf{p}_{des}) \leq \delta} - \alpha \cdot \textrm{CD}(\mathbf{p}_t, \mathbf{p}_{des}),
\end{align}
where $\delta$ is a distance threshold for success check and $\alpha$ is a constant. 
\end{itemize}
In this work, we define the loss $\mathcal{L}_{\text{RL}}$ as a combination of actor, critic, and entropy losses in \cite{haarnoja2018soft}.

Our framework updates the encoder $\Phi_\phi$ and its associated policy $\pi$ employing off-policy RL, particularly soft actor-critic~\cite{haarnoja2018soft}. At each time step $t$, we embed a state $\mathbf{s}_t$ into a latent state $(\mathbf{s}_{t}^{\text{p}}, \mathbf{z}_t)$ using the encoder $\Phi_\phi$, and then take the embedding as input for the actor network $\pi$ to determine an action~$\mathbf{a}_t$. Simultaneously, we update the critic and encoder networks by back-propagating the loss $\mathcal{L}_{\text{RL}}$ . During this update, we store transitions $(\mathbf{s}_{t}^{\text{p}}, \mathbf{p}_{t}, \mathbf{a}_t, r_t, \mathbf{s}_{t+1}^{\text{p}}, \mathbf{p}_{t+1})$ into the buffer $\mathcal{B}$, where $r_t$ denotes the output from $R(\vs_t, \va_t)$ and we use the buffer for sampling each batch $\mathcal{B}'$ for the off-policy updates.

\noindent\textbf{Contrastive learning}. To improve sample efficiency in RL, we adopt contrastive learning by adding an auxiliary task, inspired by CURL~\cite{laskin2020curl}. The task is to improve the discrimination capability of the \textit{query} encoder $\Phi_\phi$ while learning the DOM policy. The discrimination requires comparing query-key pairs to ensure that a query input $\mathbf{z}^{\text{q}}$ is close to a positive keys $\mathbf{z}^+$ and far away from negative keys $\{\mathbf{z}^{-(k)}\}_{k=1}^K$, where $K$ is the number of negative keys. However, conventional key generation often does not capture the similarity of keys in time series~\cite{lee2024soft}. To address this issue, we introduce a novel query-key selection strategy for time series and the update of the \textit{query} encoder leveraging an information noise-contrastive estimation (InfoNCE) loss~\cite{oord2018representation}. 

Consider a batch $\mathcal{B}'$ that contains sequences (i.e., episodes) of experience $[\mathcal{E}^{(1)}, ... , \mathcal{E}^{(|\mathcal{B}'|)}]$. Note that, for notational simplicity, we assume the batch $\mathcal{B}'$ is a set of point-cloud sequences $[\mathcal{E}^{(1)}_{\mathbf{p}}, ... , \mathcal{E}^{(|\mathcal{B}'|)}_{\mathbf{p}}]$. We represent the $i$-th sequence as $\mathcal{E}^{(i)}_\mathbf{p}=[\mathbf{p}_1^{(i)}, ..., \mathbf{p}_T^{(i)}]\in\mathbb{R}^{(3\times N_p) \times T}$, where $N_p$ and $T$ denote the number of points and the sequence length, respectively. We then represent the sequence embedding as $\mathcal{E}_{\mathbf{z}}^{(i)}=\Phi_{\phi}(\mathcal{E}_{\mathbf{p}}^{(i)})=[\mathbf{z}_1^{(i)}, ... , \mathbf{z}_T^{(i)} ]\in \mathbb{R}^{N_{\vz} \times T}$.

For the contrastive-loss computation, we generate query-key pairs. For the query selection, we randomly sample a point cloud $\mathbf{p}_t^{(i)}$ to obtain $\mathbf{z}^{\text{q}}=\Phi_\phi(\mathbf{p}_t^{(i)})$ as a query input from a randomly selected episode $\mathcal{E}_\mathbf{p}^{(i)}$. For the positive-key selection, we first sample the top-$M$ similar episodes from the buffer $\mathcal{B}$ with respect to the episode $\mathcal{E}_{\mathbf{p}}^{(i)}$. In this work, for computational efficiency, we determine the similarity based on the minimal start-and-goal embedding distance, $d_{\text{sg}}(\mathcal{E}_{\mathbf{p}}^{(i)}, \mathcal{E}_{\mathbf{p}}^{(j)})= \mathbf{z}_1^{(i)}\boldsymbol{\cdot} \mathbf{z}_1^{(j)}  +  \mathbf{z}_T^{(i)}\boldsymbol{\cdot} \mathbf{z}_T^{(j)} $, where `$\boldsymbol{\cdot}$' represents a dot product. We then select a point cloud $\mathbf{p}_{t'}$ in an episode with the minimum dynamic time-warping (DTW) distance from $\mathcal{E}_{\mathbf{p}}^{(i)}$ to obtain $\mathbf{z}^+=\Phi_{\phi_\text{key}}(\mathbf{p}_{t'})$, where $\Phi_{\phi_{\text{key}}}$ is the \textit{key} encoder~\cite{he2020momentum}, initialized with $\phi$. Note that $t'$ is a time step matched to the time step $t$ of $\mathcal{E}_{\mathbf{p}}^{(i)}$ by DTW. For the negative key selections, we randomly sample the $K$ number of point clouds from $\mathcal{B}'\backslash\{\mathcal{E}_{\mathbf{p}}^{(i)}\}$ and embed them as $\{\mathbf{z}^{-(k)}\}_{k=1}^{K}$ using $\Phi_{\phi_{\text{key}}}$.

Given the query-key pairs, we compute the InfoNCE loss $\mathcal{L}_{\text{infoNCE}}$ to optimize the encoders $\Phi_\phi$ and $\Phi_{\phi_{\text{key}}}$:
\begin{equation}
\log \frac{\exp(\mathbf{z}^\text{q}\boldsymbol{\cdot}\mathbf{z}^+/\tau)}{\exp(\mathbf{z}^\text{q}\boldsymbol{\cdot}\mathbf{z}^+/\tau) + \sum_{k=1}^{K}\exp(\mathbf{z}^\text{q}\boldsymbol{\cdot} {\mathbf{z}^{-(k)}}/\tau)}
\end{equation}
where $\tau$ is a temperature parameter ($\in \mathbb{R}^+$). As MoCO~\cite{he2020momentum}, we update $\phi_{\text{key}} = m \phi_{\text{key}} + (1 - m) \phi   $ using the exponential moving average (EMA) method, where $m$ is the momentum coefficient ($\in[0,1)$). The encoder is then able to capture subtle distinctions between object configurations, which is critical for manipulation tasks involving deformable objects. In this work, we set \( \tau = 0.1\).

In summary, the fine-tuning stage of INR-DOM, inspired by the CURL framework, involves refining the latent space using contrastive learning to capture task-relevant features while maintaining generalization. This enhanced representation, coupled with RL, enables the model to learn effective manipulation policies more efficiently.
\section{Experimental Setup}
\label{sec:experiment}
Our experimental evaluation aims to answer two key questions: 1) Does the proposed representation provide consistent and occlusion-robust state information for DOM? 2) Further, does the proposed method improve the effectiveness of DOM in the real world? We perform quantitative evaluations through simulation and qualitative studies with a real robot.

\subsection{Quantitative Evaluation through Simulation}
\label{ssec:quantitative}
We statistically evaluate the capabilities of \textbf{occlusion recovery} and \textbf{task completion} in INR-DOM. 

\begin{figure}[t]
\centering
    \includegraphics[width=\columnwidth]{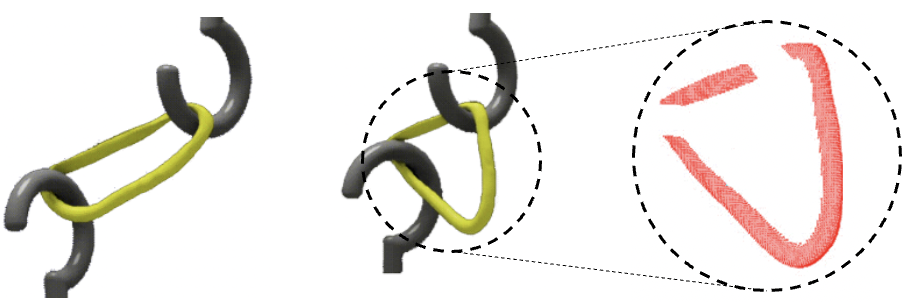}
    \caption{\label{fig:oring_collection_2} Examples of randomly twisted and stretched rubber bands in simulation. The red points represent partial point clouds.}
\end{figure}

\noindent\textbf{Occlusion recovery}: We assess the occlusion-tolerant reconstruction capability of INR-DOM through IsaacSim, a physics simulator from NVIDIA~\cite{NVIDIA_IsaacSim_2025}. We build a random manipulation dataset for nine types of circular rubber bands. Each band has a pair of \textit{inside diameter} (ID) $d_{\text{ID}}$ and \textit{cross-sectional diameter} (CSD) $d_{\text{CSD}}$, where $\{(d_{\text{ID}}, d_{\text{CSD}}) \mid d_{\text{ID}} \in \{\SI{6}{\cm}, \SI{10}{\cm}, \SI{14}{\cm}\}, d_{\text{CSD}} \in \{\SI{1}{\cm}, \SI{2}{\cm}, \SI{3}{\cm}\} \}$. As shown in Fig.~\ref{fig:oring_collection_2}, by randomly twisting and stretching them, we collect a total of $90,000$ pairs of partial and complete point clouds, $(\mathbf{p}, \bar{\mathbf{p}})$, $10,000$ pairs each. Note that we restrict the randomization to a maximum of one twist in each direction, and a maximum stretch of twice the original length. 

Given the nine classes of dataset, we perform a leave-one-out cross-validation that measures how representation models reconstruct the full geometry of rubber bands from a partial view of point clouds. During the pre-training stage of INR-DOM, we randomly sample $1,024$ points from each of the areas, on, near and off surfaces of the target object to compute $\mathcal{L}_{\text{SDF}}$ and $\mathcal{L}_{\text{skel}}$. To define the medial-axis points, we sample and track the $128$ nodes spaced equally along the band and closest to the center of the cross section.

For comparison, we employ three baseline methods:
\begin{itemize}[leftmargin=*]
\item PCN~\cite{yuan2018pcn}: Point completion network; a shape-completion network with folding-based decoding that generates a dense and complete point cloud given a partial point cloud as input.
\item PointTr~\cite{yu2021pointr}: A transformer-based encoder-decoder for point-cloud completion.
\item Point2Vec~\cite{zeid2023point2vec}: A student-teacher framework of latent embedding and completion network.
\end{itemize}
After reconstructing the full geometry, we compute the earth mover's distance (EMD) and Chamfer distance (CD) between the reconstructed and complete point clouds. To handle SDF, we convert it into a mesh-based point cloud using the Marching Cubes algorithm~\cite{10.1145/37402.37422}.

\begin{figure}[t]
    \includegraphics[width=\columnwidth]{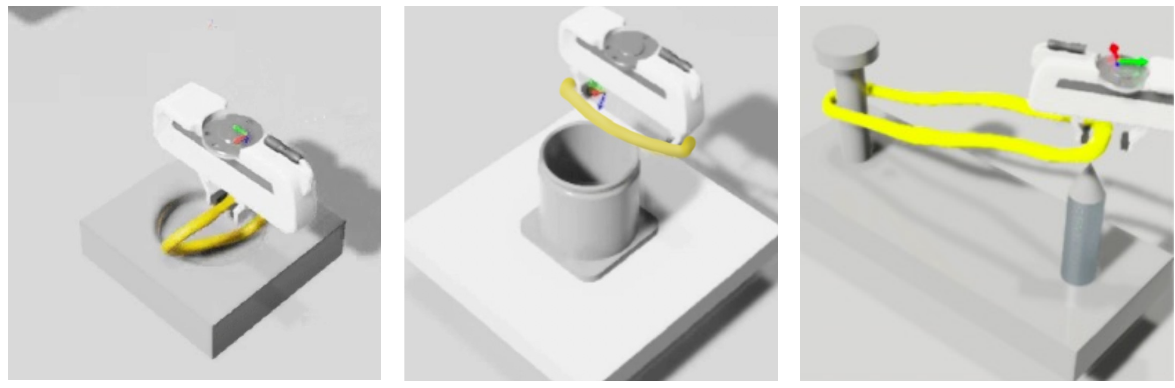}
    \caption{Examples of three deformable-object manipulation environments: \textit{sealing}, \textit{installation}, and \textit{disentanglement}.}
    \label{fig:oring_manipulation}
\end{figure}

\noindent\textbf{Manipulation tasks}: We evaluate the effectiveness of fine-tuned representations in DOM by building three tasks: \textit{sealing}, \textit{installation}, and \textit{disentanglement} (see Fig.~\ref{fig:oring_manipulation}). For each task, we use a simulated parallel-jaw gripper with an RGB-D camera to manipulate a rubber band or an O-ring to achieve a desired state. We describe the task setup below.
\begin{itemize}[leftmargin=*]
\item \textit{Sealing}: A gripper seals a groove by inserting a rubber band after grasping a randomly placed band. To enhance sampling efficiency, we reduce the action space to $(\mathbf{a}^{\text{lin}}_t, \mathbf{a}^{\text{gripper}}_t)\in\mathbb{R}^4$ setting $\mathbf{a}^{\text{ang}}_t$ to zero. 
\item \textit{Installation}: A gripper installs an O-ring onto the circular groove of a cylinder after grasping a randomly placed O-ring. The action space is $(\mathbf{a}^{\text{lin}.xz}_t, \mathbf{a}^{\text{ang}.y}_t , \mathbf{a}^{\text{gripper}}_t)\in\mathbb{R}^4$, where $\mathbf{a}^{\text{lin}.xz}_t$ and $\mathbf{a}^{\text{ang}.y}_t$ represent a two-dimensional velocity perpendicular to the direction of the parallel jaw's open/close axis and an angular velocity along the same axis, respectively.
\item \textit{Disentanglement}: A gripper untangles an entangled rubber band between two upright poles by grasping and repositioning it onto the poles. We initialize a band between two poles with up to two random entanglements in either of the two directions. The action space is $(\mathbf{a}^{\text{lin}}_t, \mathbf{a}^{\text{ang}.z}_t , \mathbf{a}^{\text{gripper}}_t)\in\mathbb{R}^5$, where $\mathbf{a}^{\text{ang}.z}_t$ represents an angular velocity constrained along the vertical axis.
\end{itemize}
For all tasks, we reduce the space of the partial point clouds to $\mathcal{P} \subset \mathbb{R}^{3\times1024}$, through the iterative farthest point sampling of the input point cloud. 

For training, we fine-tune INR-DOM with all types of rubber bands from the occlusion recovery study and update its policy network using an off-policy RL method, that is, SAC~\cite{haarnoja2018soft}. The policy network is a multilayer perceptron with three hidden layers consisting of $256$, $128$, and $64$ nodes, respectively, each using ReLU activation. For testing, we assess the task completion performance of INR-DOM in $100$ environments, including $30$ randomly generated and configured rubber bands. The ID and CSD of the bands range from $[\SI{6}{\cm}, \SI{14}{\cm}]$ and $[\SI{1}{\cm}, \SI{3}{\cm}]$, respectively.

For comparison, we employ six baseline methods:
\begin{itemize}[leftmargin=*]
\item \textbf{PCN+SAC}, \textbf{PoinTr+SAC}, and \textbf{Point2Vec+SAC}: RL-based DOM methods trained using the latent representations introduced in the occlusion-recovery study.
\item \textbf{CFM}~\cite{yan2021learning}: A model predictive control framework that leverages latent representations and dynamics models learned with contrastive estimation for DOM.
\item \textbf{DeformerNet}~\cite{thach2022learning}: A visual servoing method that minimizes the difference between the current and target point clouds of the deformable object.
\item \textbf{ACID}~\cite{shen2022acid}: A model-based planning method that calculates action costs based on the difference between the learned implicit representations.
\end{itemize}
We evaluate our method with these baselines over 100 trials, each with a maximum of $200$ time steps.

\subsection{Qualitative Evaluation}
\label{ssec:qualitative}

We demonstrate the fine-tuning and testing of our method in three real-world DOM environments using a Franka Emika Panda arm, as shown in Fig.~\ref{fig:main}. These environments are similar to those used in the quantitative evaluation. Each setup includes a rubber band with specific dimensions: 1) $d_{\text{ID}}=\SI{10}{\cm}$ and $d_{\text{CSD}}=\SI{0.6}{\cm}$ for \textit{sealing}, 2) $d_{\text{ID}}=\SI{6}{\cm}$ and $d_{\text{CSD}}=\SI{1}{\cm}$ for \textit{installation}, and 3) $d_{\text{ID}}=\SI{20}{\cm}$ and $d_{\text{CSD}}=\SI{1}{\cm}$ for \textit{disentanglement}. To observe the band, we mount an Orbbec Femto Bolt RGB-D camera on the table capturing the point cloud $\mathbf{p}_t$ of the band by segmenting and tracking it with a pre-trained segmentation anything model 2 (SAM2)~\cite{anonymous2024sam}. For more effective localization and grasping of the band, we employ two RealSense D405 RGB cameras mounted on the wrist, extending the latent state $\mathbf{s}'_t$ with a concatenated vector of RGB features processed by ResNet10~\cite{gong2022resnet10}. In this work, we acquire the segmented points $\mathbf{p}_t$ after outlier removal at \SI{10}{\hertz} along with two RGB observations captured at \SI{15}{\hertz}. 

Unlike simulation, we design an image-based reward classifier that assigns a high reward when the current RGB image closely matches the final scene from expert demonstrations; otherwise, it assigns a zero reward. To train this classifier, we collect $20$ demonstrations by teleoperating the real robot using a 3-dimensional space mouse. We use these demonstrations to populate the replay buffer during the fine-tuning stage. Further, we adopt a sample-efficient robotic reinforcement learning (SERL) framework~\cite{DBLP:conf/icra/LuoH0TBSSF0L24}, which incorporates RL with prior data (RLPD)~\cite{ball2023efficient}. For RLPD, we construct training batches with half of the samples drawn from the demonstrations and the other half from the online replay buffer. We use 20 successful episodes as demonstrations for the prior data. Note that we set the update-to-data (UTD) ratio to $4$. This allows leveraging human demonstrations to accelerate the learning process and minimize the sim-to-real gap in representation.

The robot operates with a Cartesian impedance controller running at \SI{1}{\kilo\hertz} while the policy network runs at \SI{10}{\hertz}. We perform all training on a Threadripper Pro 5975WX and an NVIDIA RTX A6000 GPU.

\begin{figure}[t]
\centering
    \includegraphics[width=\columnwidth]{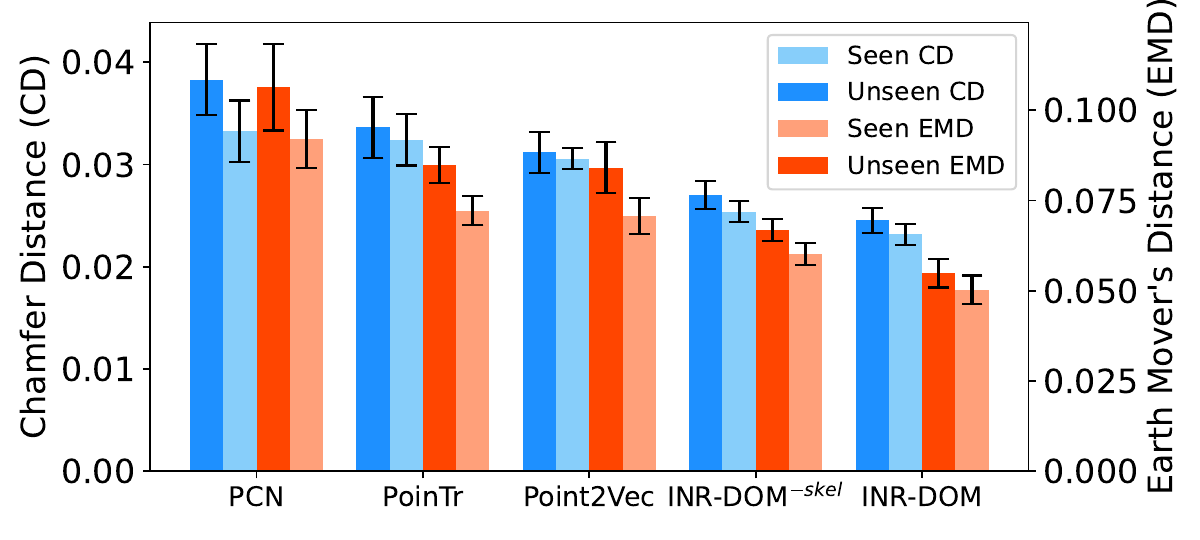}
    \caption{Comparison of point-cloud reconstruction performance for both seen and unseen types of partially observable rubber bands. The blue and red bars represent the reconstruction errors measured by chamfer distance (CD) and earth mover's distance (EMD), respectively. Note that INR-DOM$^{-skel}$ refers to a variant of INR-DOM that was not pre-trained using the $\mathcal{L}_\text{skel}$ loss.    
    }
    \label{fig:oring_disentangle}
\end{figure}

\begin{figure}[t]
\centering
    \includegraphics[width=\columnwidth]{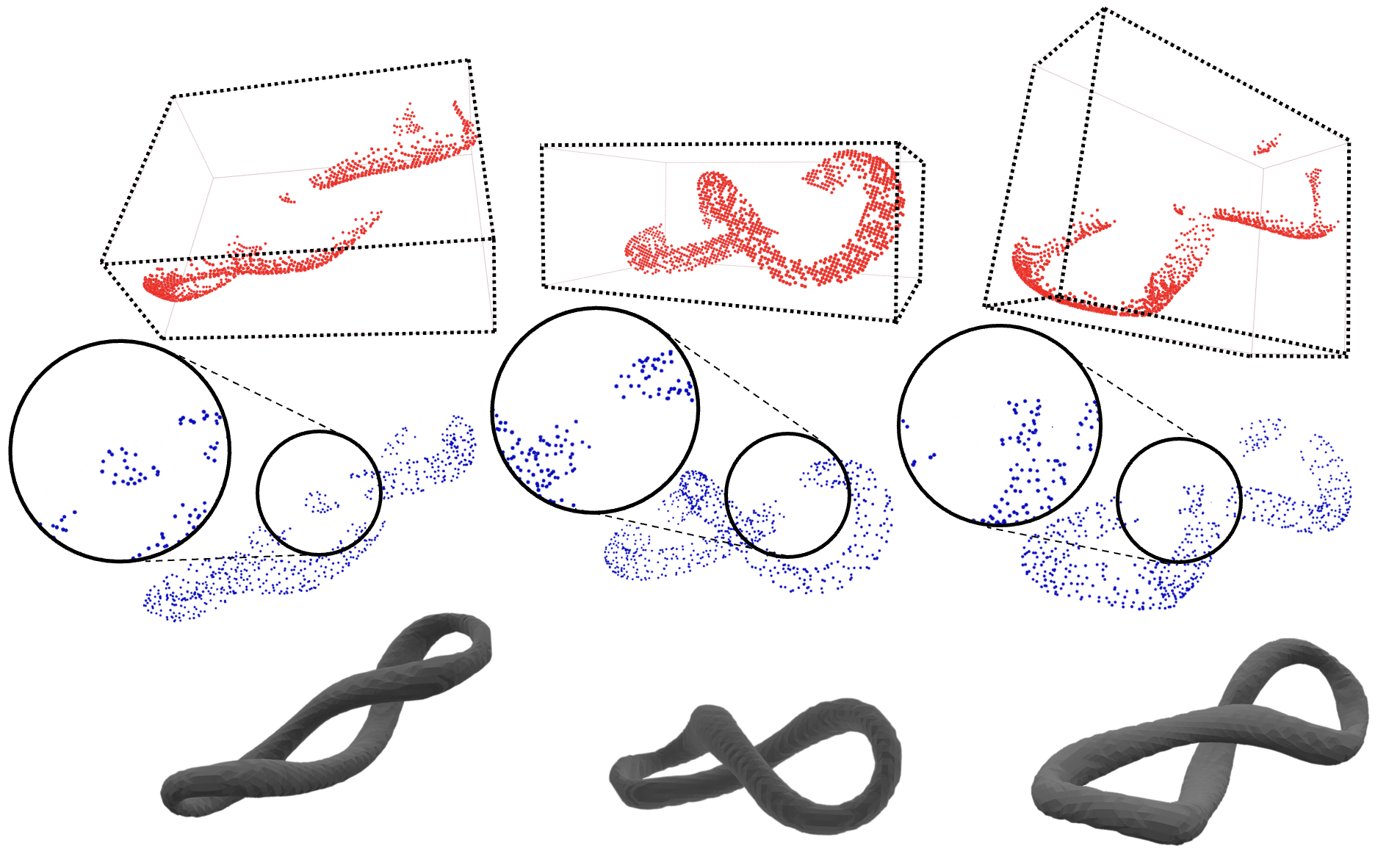}
    \caption{\label{fig:recon_qual}  Comparison of occlusion-robust reconstruction performance between INR-DOM and Point2Vec. (\textbf{Top}) Point-cloud inputs of partially observable elastic bands. (\textbf{Middle}) Point clouds reconstructed by Point2Vec. The larger circles highlight magnified views of regions where reconstruction failed, indicated by smaller circles. (\textbf{Bottom}) SDF-based Meshes from INR-DOM.
    }
\end{figure}

\begin{table}[t]
\centering
\caption{Comparison of task success rates [\%] across three simulated environments, based on the evaluation of 100 trials per environment. The superscripts, \textit{-p} and \textit{-cl}, indicate versions of the target method that were not trained with pre-training and contrastive learning, respectively.
}
\label{tab:dom_performance}
\begin{tabular}{lccc}
\toprule
\textbf{Model}       & \textit{Sealing}& \textit{Installation}& \textit{Disentanglement}\\
\midrule
PCN\cite{yuan2018pcn} + SAC\cite{haarnoja2018soft}& 13                    & 38    &6                                       \\
PoinTr\cite{yu2021pointr} + SAC\cite{haarnoja2018soft}& 23                    & 38 &14                                          \\
Point2Vec\cite{zeid2023point2vec} + SAC\cite{haarnoja2018soft}& 14                    & 40            &22                               \\
Point2Vec$^{-p}$\cite{zeid2023point2vec} + SAC\cite{haarnoja2018soft}& 2                    & 1          &1                                \\
CFM\cite{yan2021learning} + SAC\cite{haarnoja2018soft}& 41                    & 47                         &23                 \\
DeformerNet\cite{thach2022learning}          & 41                    & 53                                   &19       \\
ACID\cite{shen2022acid}                 & 44                    & 58                                      &26    \\
INR-DOM$^{-p}$& 20& 29 &16\\
 INR-DOM$^{-cl}$& 58& 61 &54\\
\textbf{INR-DOM}& \textbf{85}       & \textbf{89} & \textbf{75}                    \\
\bottomrule
\end{tabular}
\end{table}

\begin{figure}[t]
\centering
    \subfloat[Distribution of 10 trajectory embeddings from the \textit{disentanglement} task]{\includegraphics[width=\columnwidth]{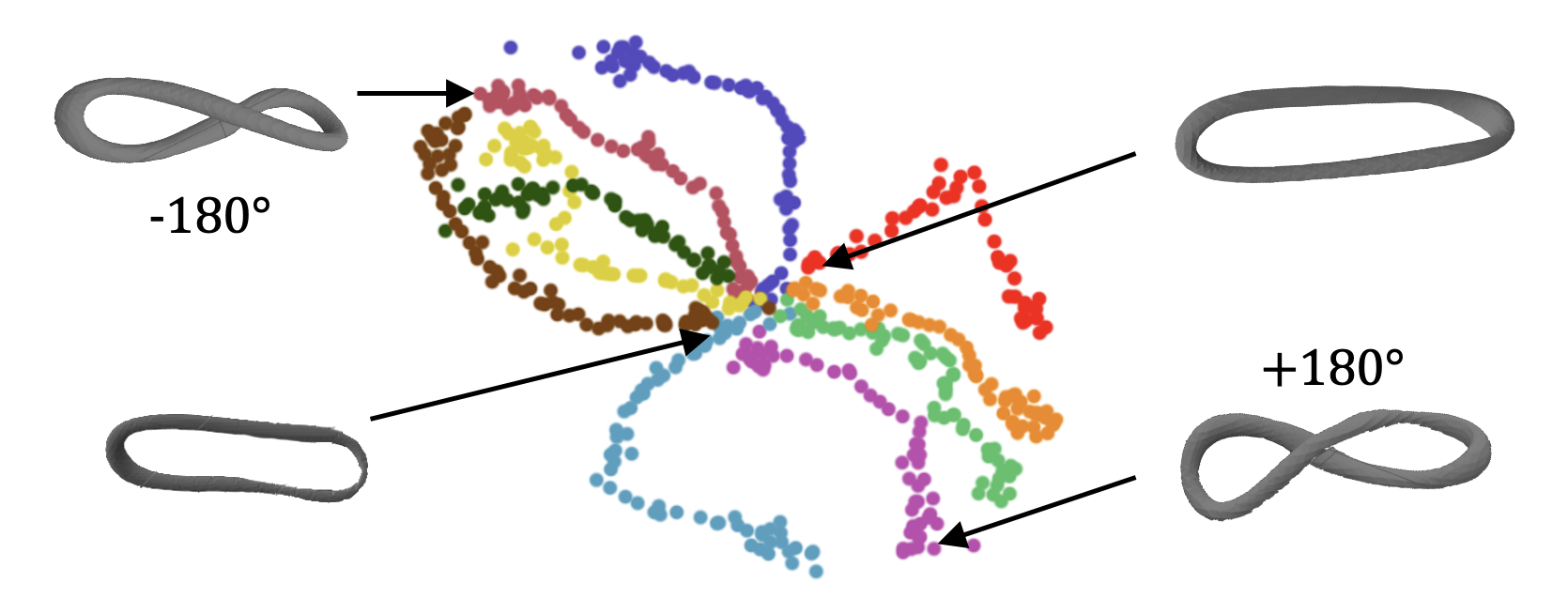}}\label{fig:tsne_a}
    \hfill
    \subfloat[Distribution of $2\cdot 10^4$ embeddings from random manipulation dataset used in the pre-training stage]{\includegraphics[width=\columnwidth]{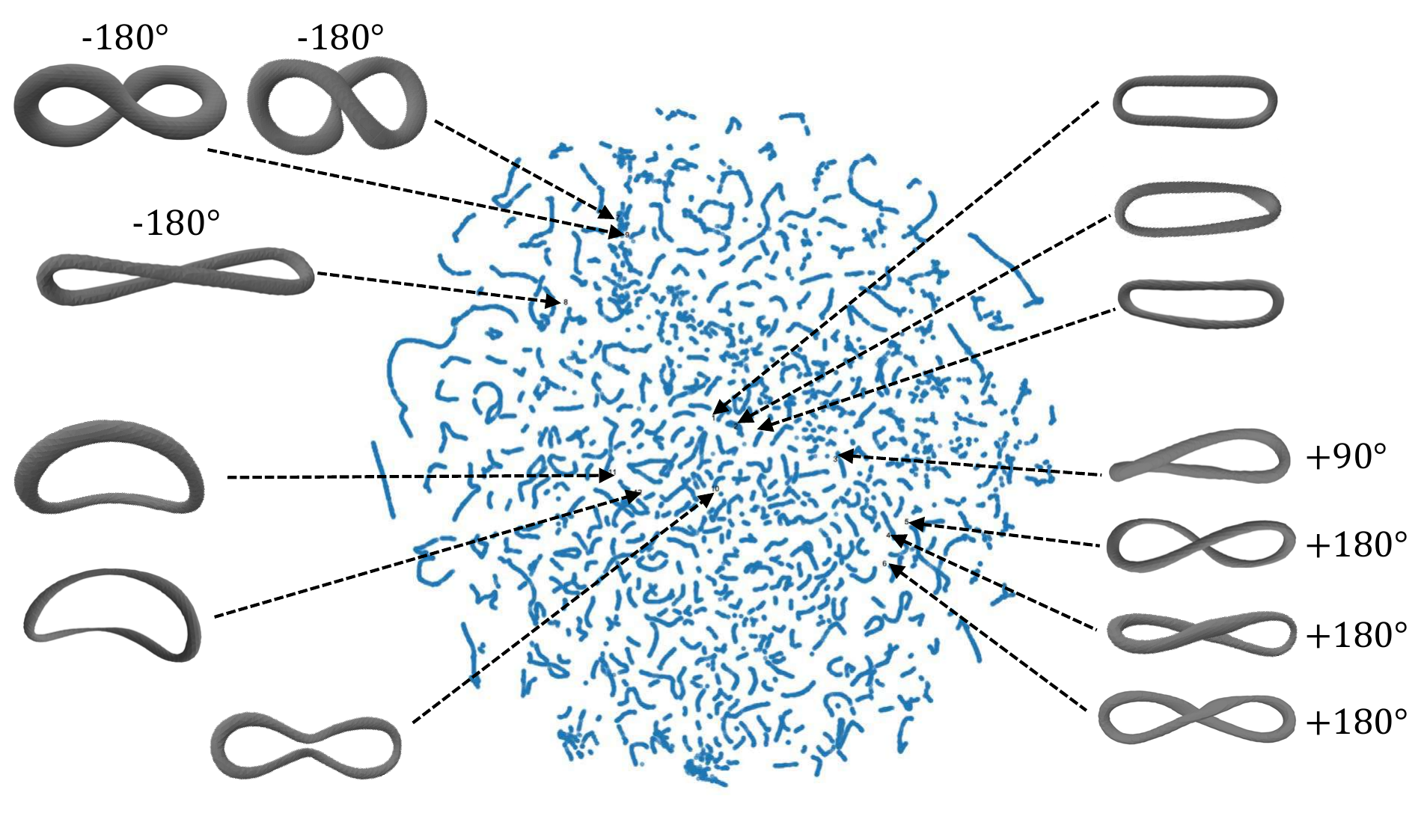}}
    \label{fig:tsne_b}
    \caption{t-SNE visualization of latent state vectors $\mathbf{z}$ from the fine-tuned encoder $\Phi_\phi$. We also visualize rubber band meshes corresponding to selected points to show how the latent state effectively captures the entanglements of the band. The signed numbers adjacent to the meshes represent twist angles and directions. 
    }\label{fig:tsne} 
\end{figure}

\section{Evaluation Results}
\subsection{Evaluation through Simulation Studies}
We statistically evaluate the occlusion-robust reconstruction capabilities of pre-trained INR-DOM on both seen and unseen partially observable bands. INR-DOM shows the lowest reconstruction errors using CD, even with previously unseen objects, as illustrated in Fig.~\ref{fig:oring_disentangle}. Remarkably, INR-DOM outperforms baselines on unseen objects more effectively than baselines do on seen objects. Further, INR-DOM achieves superior performance without structural prior $\mathcal{L}_{\text{skel}}$, exceeding all baselines. In addition, we assess performance using EMD, which better accounts for global distribution differences crucial for compressed or stretched DOs. The results with EMD align with those of CD, confirming that INR-DOM provides robust representations for tasks involving partially observable DOs.

To show the reconstruction quality, we present examples from INR-DOM and Point2Vec in Fig.~\ref{fig:recon_qual}. INR-DOM accurately reconstructs complete geometry from any observation angle, whereas Point2Vec struggles with occluded, particularly intertwined, parts. This highlights the effectiveness of our pre-training method for DOs.

Next, we assess the consistent representation capabilities of INR-DOM in the \textit{disentanglement} task. Fig.~\ref{fig:tsne} presents t-SNE visualizations of latent state vectors from the fine-tuned INR-DOM encoder $\Phi_\phi$, showing consistent mapping of similar configurations to adjacent regions and clear distinction of intertwined state directions. For example, in Fig.~\ref{fig:tsne} (a), all disentanglement trajectories converge to the center region as the task completes, with $+180^\circ$ and $-180^\circ$ twists placed on opposite sides. This structured latent space allows the RL agent to recognize subtle yet crucial variations, such as twist counts, thereby improving success rates.

We evaluate the task-relevant representation capabilities of INR-DOM through three simulated DOM tasks. Table~\ref{tab:dom_performance} shows INR-DOM outperforms all other methods, achieving the highest task-success rates, an average rate $40.3\%$ higher than ACID, the next best approach. INR-DOM exhibits superior performance in the challenging \textit{disentanglement} task, characterized by stretched and randomly intertwined states. Notably, INR-DOM$^{-cl}$, fine-tuned only via RL, ranks second, demonstrating the efficacy of contrastive learning in distinguishing state spaces for DOM. In contrast, INR-DOM$^{-p}$, which lacked pre-training, does not show performance gains, highlighting that the reconstruction-based pre-training is crucial for effective initialization of state-representation learning. In contrast, other recent DOM frameworks underperform, primarily due to their inability to capture local structures and their non-rigid transformations in DOs.

Lastly, we analyze the lower performance of point-cloud completion approaches, such as PCN, PointTr, and Point2Vec. Their embeddings capture overall structure but struggle with continuous and fine deformations due to discrete sampling. In contrast, our SDF loss with gradient regularization and alignment not only allows adaptive resolution but also ensures high accuracy in surface completion. Although the latest approach, Point2Vec, successfully encodes local patches with contextual information, our target objects lack explicit contextual patches and instead require accurate positional and structural details. Due to its patch normalization, Point2Vec loses such information, resulting in collapsed or overlapped geometries.

\begin{figure}[t]
\centering
    \includegraphics[width=\columnwidth]{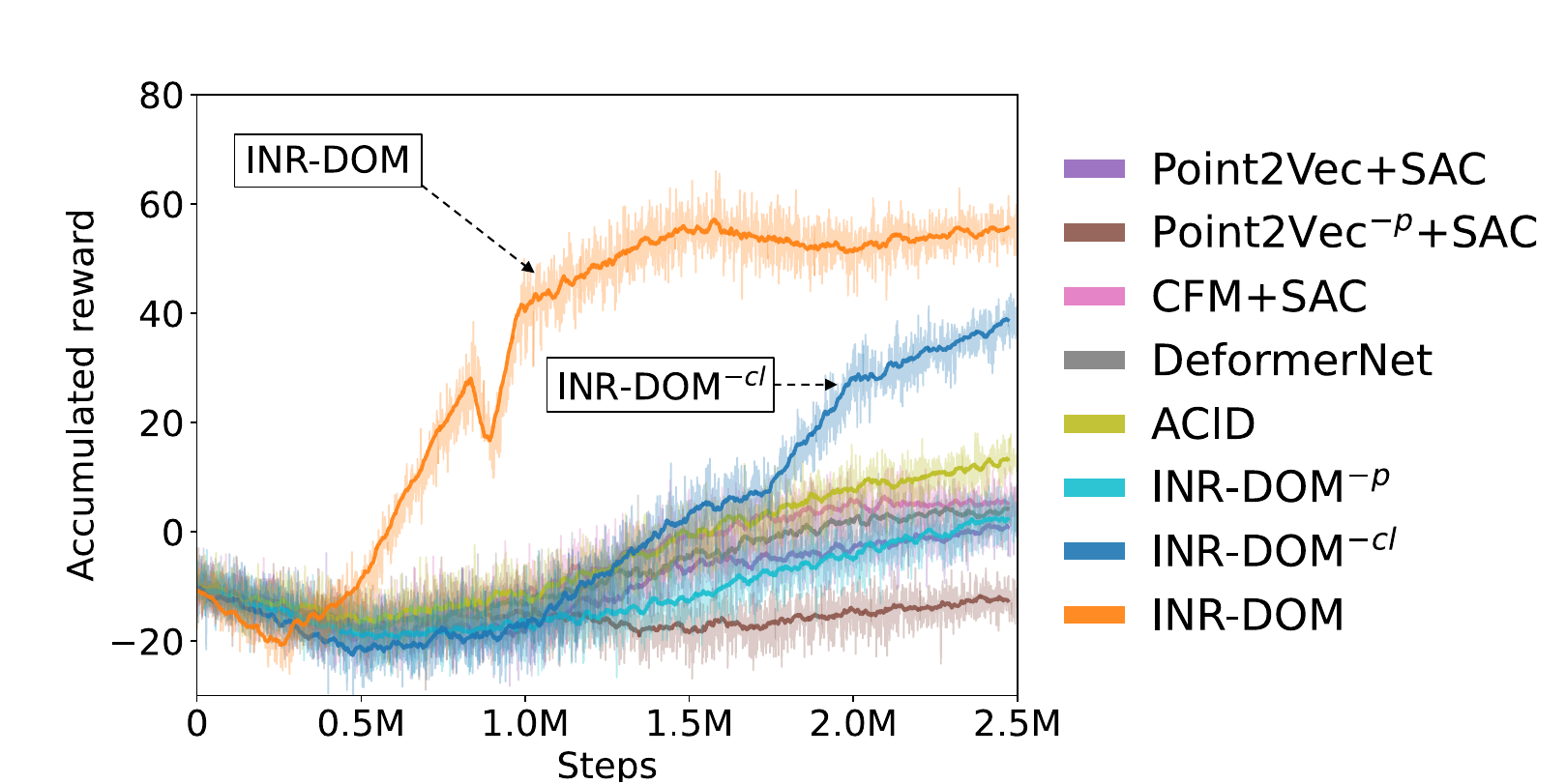}
    \caption{Comparison of accumulated reward curves during training between INR-DOM and baseline models.
    }\label{fig:reward_curves}
\end{figure}

\begin{figure*}[t]
\centering
    \includegraphics[width=1.0\textwidth]{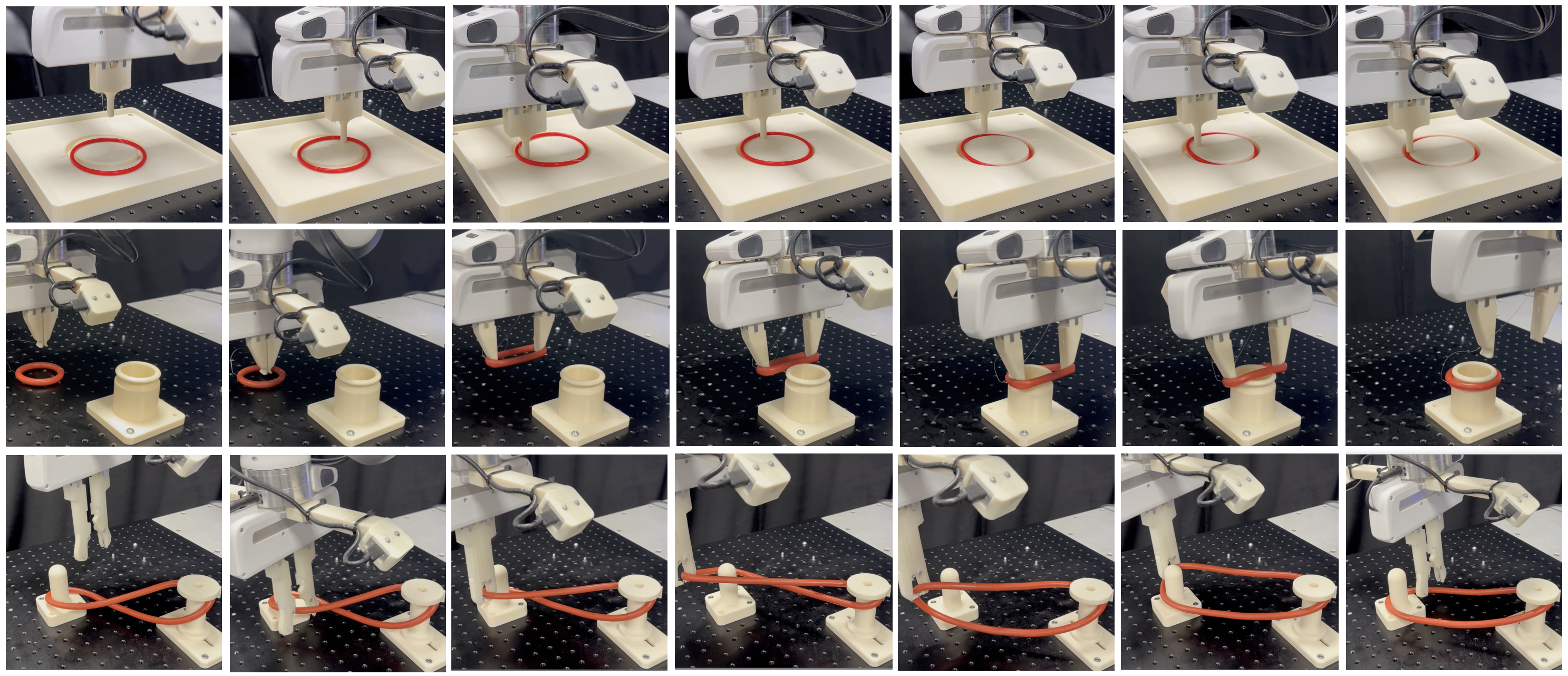}
    \caption{\label{fig:untwist_real} Demonstrations of INR-DOM's manipulation capability in the three tasks. (\textbf{Top}) In \textit{sealing} task, INR-DOM learns to drag and insert a rubber band into a groove. (\textbf{Middle}) In \textit{installation} task, INR-DOM learns to install an O-ring onto a cylinder's groove.
(\textbf{Bottom}) \textit{disentanglement} task, INR-DOM learns to disentangle the band between two poles by precisely identifying the twisted status.       
}
\end{figure*}

\subsection{Learning Curve Analysis}
Fig.~\ref{fig:reward_curves} shows the comparison of the accumulated reward curves during training for the \textit{disentanglement} task between INR-DOM with variants and baseline methods. INR-DOM shows superior learning efficiency, achieving the highest accumulated rewards compared to all baselines. INR-DOM$^{-cl}$ particularly ranks second in terms of the accumulated rewards at the end, though it lags significantly behind INR-DOM. This underscores that contrastive learning enhances effective exploration in RL by fostering exploratory representations. In contrast, INR-DOM$^{-p}$ achieves only a $15\%$ success rate after $2.5$ million training steps, indicating that exploratory representations are challenging to enhance over steps without a well-structured state space.  

\subsection{Evaluations in Real-World Settings}
Finally, we demonstrate the real-world applicability of INR-DOM across three DOM tasks. Fig.~\ref{fig:untwist_real} (Top) displays the Panda arm with a 3D-printed tip sealing a groove by dragging and inserting a randomly placed red rubber band, effectively targeting the protruding parts for pressing actions. Fig.~\ref{fig:untwist_real} (Middle) shows the robot installing an O-ring onto a cylinder's groove, opening the ring, and leveraging one-sided contact for insertion. Fig.~\ref{fig:untwist_real} (Bottom) illustrates the Panda arm with a long parallel-jaw gripper, tip-grasping and disentangling a randomly intertwined rubber band between two poles. INR-DOM successfully distinguishes the twisted direction as well as effectively manipulates the arm, achieving a $90\%$ task-success rate in each task over ten trials with various initial conditions. This robust performance, coupled with successful learning from the fine-tuning of $20$ episodes, confirms the applicability of INR-DOM to various real-world DOM tasks.

\begin{figure}[h]
    \centering
    \includegraphics[width=0.48\textwidth]{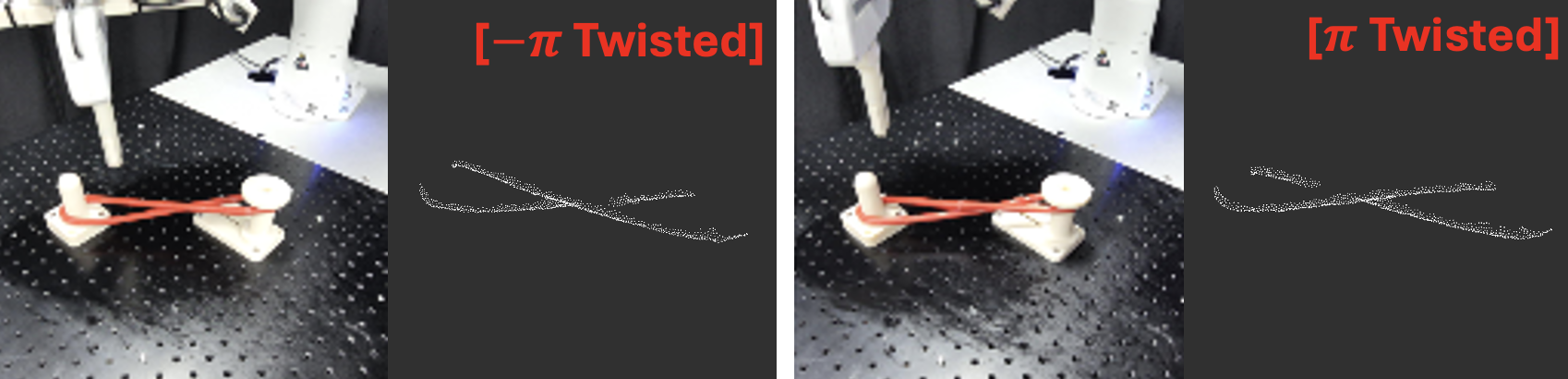}
    \caption{Visual ambiguity in RGB vs. point cloud observations for $-\pi$ and $\pi$ twist configurations.}
    \label{fig:rgb_vs_pcd_caption}
\end{figure}

In addition, we compare INR-DOM with SERL, a representative image-based manipulation method on the real-world \textit{disentanglement} task. To increase task difficulty, we extend the pole distance to stretch the bands, resulting in tighter intersections. SERL uses an RGB camera, whereas our method relies on a depth camera. As shown in Table~\ref{tab:rgb_vs_pcd}, INR-DOM successfully distinguishes between $\pm$\SI{180}{\degree} twist directions, while SERL fails due to visual ambiguities at the intersecting bands (see Fig.~\ref{fig:rgb_vs_pcd_caption}).
\begin{table}[h]
\centering
\caption{Comparison of task success rates [\%] between an SERL image-based method and our method in the \textit{disentanglement} task over $20$ trials. }
\label{tab:rgb_vs_pcd}
\begin{tabular}{lc|lc}
\toprule
\textbf{Model} & Rate [\%] & \textbf{Model} & Rate [\%] \\
\midrule
SERL & 55\% & INR-DOM & 80\% \\
\bottomrule
\end{tabular}
\end{table}

\section{Conclusion}

We introduced an implicit neural representation learning method for elastic deformable object manipulation (INR-DOM). INR-DOM models consistent and occlusion-robust state representations associated with partially observable elastic objects by learning to reconstruct a complete and implicit surface represented as a signed distance function. To obtain task-relevant state representations and manipulation policy, INR-DOM fine-tunes its internal encoder to have exploratory representations through RL-based contrastive learning. The statistical evaluation shows that our method outperforms state-of-the-art baselines in terms of reconstruction error and task success rate on novel objects and manipulation settings. We successfully demonstrate the proposed INR-DOM transfer into real-world DOM tasks.
\section{Limitations}
\begin{itemize}[leftmargin=*]
\item\textbf{Deformable linear object}: Our studies focus on evaluating deformable linear objects, such as rubber bands, though the usage is not limited to specific shapes. Although skeleton loss $\mathcal{L}_{skel}$ may reduce its applicability, 3D skeletonization is generalizable to any volumetric object. Extending this method to a variety of deformable object types remains a goal for future research.

\item\textbf{Segmented observation}: The performance of INR-DOM depends on the accuracy of the point-cloud segmentation model. This dependency makes INR-DOM vulnerable to errors in the segmentation process, which affects the robustness and reliability of the task outcomes. Future improvements should involve processing the entire depth image and automatically focusing attention on the target deformable object. 

\item\textbf{Single task}: The fine-tuning process relies on a specific manipulation task which may limit the model's applicability to other scenarios. Moreover, INR-DOM's design focuses exclusively on a single deformable object manipulation and does not account for scenarios with multiple interacting objects, common in complex environments. However, the architecture does not inherently restrict itself to single-task learning. The adoption of multi-task reinforcement learning~\cite{10892354} or few-shot policy generalization~\cite{xu2022prompting} extends beyond the scope of this work. Expanding to multitask RL approaches is our next objective.

\item\textbf{Dynamic interaction}: This work does not extensively investigate the dynamics of interaction between the robot and deformable, particularly stretchable objects. The manipulation tasks focus primarily on the final outcomes of actions, rather than on modeling the interactive dynamics essential for tasks that require continuous adjustments based on real-time feedback. This limitation may restrict the applicability of INR-DOM to more interactive and dynamic tasks, where understanding and adapting to ongoing changes during manipulation are crucial.
\end{itemize}

\section*{Acknowledgments}
This work was partly supported by Institute of Information \& communications Technology Planning \& Evaluation (IITP) grants funded by the Korea government (MSIT) (No. RS-2022-II220311, RS-2024-00509279, and RS-2024-00336738) and Samsung Electronics Co., Ltd, South Korea (No. IO220811-01961-01).


\bibliographystyle{plainnat}
\bibliography{references}

\end{document}